# Towards Multi-agent Reinforcement Learning for Wireless Network Protocol Synthesis

Hrishikesh Dutta and Subir Biswas
*Electrical and Computer Engineering, Michigan State University, East Lansing, MI, USA*
*duttahr1@msu.edu, sbiswas@msu.edu*

*Abstract* — This paper proposes a multi-agent reinforcement learning based medium access framework for wireless networks. The access problem is formulated as a Markov Decision Process (MDP), and solved using reinforcement learning with every network node acting as a distributed learning agent. The solution components are developed step by step, starting from a single-node access scenario in which a node agent incrementally learns to control MAC layer packet loads for reining in self-collisions. The strategy is then scaled up for multi-node fully-connected scenarios by using more elaborate reward structures. It also demonstrates preliminary feasibility for more general partially connected topologies. It is shown that by learning to adjust MAC layer transmission probabilities, the protocol is not only able to attain theoretical maximum throughput at an optimal load, but unlike classical approaches, it can also retain that maximum throughput at higher loading conditions. Additionally, the mechanism is agnostic to heterogeneous loading while preserving that feature. It is also shown that access priorities of the protocol across nodes can be parametrically adjusted. Finally, it is also shown that the online learning feature of reinforcement learning is able to make the protocol adapt to time-varying loading conditions

## I. Introduction

The objective of this paper is to explore an online learning paradigm for medium access control (MAC) in wireless networks with multiple levels of heterogeneity. Towards the long-term goal of developing a general-purpose learning framework, in this particular paper we start with a basic scenario in which wireless nodes run a rudimentary MAC logic without relying on carrier sensing and other complex features from the underlying hardware. In other words, the developed mechanisms are suitable for very simple transceivers that are found in low-cost wireless sensors, Internet of Things (IoTs), and other embedded devices.

The current best practice for programming MAC logic in an embedded wireless node is to implement known protocols such as ALOHA, Slotted-ALOHA, CSMA, and CSMA-CA (i.e., WiFi, BT, etc.) depending on the available lower layer hardware support. The choice of such protocols is often driven by heuristics and past experience of network designers. While such methods provide a standard method for network and protocol deployment, they do not necessarily maximize the MAC layer performance in a fair manner, especially in the presence of network and data heterogeneity. An example is when nodes without carrier sensing abilities run AOLHA family of protocols, their performance start degrading due to collisions when the application traffic load in the network exceeds an optimal level. Such problems are further compounded in the presence of various forms of heterogeneity in terms of traffic load, topology, and node-specific access priorities. The key reason for such performance gap is that the nodes are statically programmed with a protocol logic that is not cognizant of time-varying load situations and such heterogeneities. The proposed framework allows wireless nodes to learn how to detect such conditions and change transmission strategies in real-time for maximizing the network performance, even under node-specific access prioritization.

The key concept in this work is to model the MAC layer logic as a Markov Decision Process (MDP) [1], and solve it dynamically using Reinforcement Learning (RL) as a temporal difference solution [2] under varying traffic and network conditions. An MDP solution is the correct set of transmission actions taken by the network nodes, which act as the MDP agents. RL provides opportunities for the nodes to learn on the fly without the need of any prior training data. The proposed framework allows provisions for heterogenous traffic load, network topology, and node-specific priority while striking the right balance between node level and network level performance. Learning adjustments to temporal variations of such heterogeneities and access priorities are also supported by leveraging the inherent real-time adaptability of Reinforcement Learning. It is shown that the nodes can learn to self-regulate collisions in order to attain theoretically maximum MAC level performance at optimal loading conditions. For higher load, the nodes learn to adjust their transmit probabilities in order to reduce the effective load, maintain low levels of collisions, thus maintaining the maximum MAC level performance.

Specific contributions of this work are as follows. First, the MAC layer logic without carrier sensing is modeled as a Markov Decision Process (MDP), which is solved using Reinforcement Learning (RL). Second, it is shown that such learning has the abilities to make the agents/nodes self-regulate their individual traffic loads in order to attain and maintain the theoretically maximum throughput even when the application level traffic load is increased beyond the point of optimality. Third, the learning mechanism is enhanced to handle heterogeneous traffic load. Fourth, node-level access priorities are incorporated in the native learning process. Finally, it is preliminarily demonstrated that the core concept of MDP formulation and RL solution can be extended for non-fully connected network topologies.

## II. Network and Traffic Model

The first network model is a multi-point to point sparse network. As shown in Fig. 1: a, *N* fully connected sensor nodes send data to a wirelessly connected base station using fixed size packets. Performance, which is node- and network-level throughputs, is affected by collisions at the base station caused by overlapping transmissions from multiple nodes. The primary objective of a learning-based MAC protocol is to learn the optimal transmission strategy that can minimize collisions at the base station. It is assumed that nodes do not have carrier sensing abilities, and each of them is able to receive transmissions from all other nodes in the network.

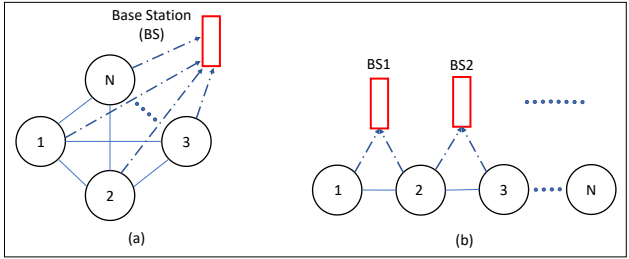

Fig. 1: Network and data upload models for (a) Fully connected nodes uploading data to a single Base Station (b) Partially connected nodes uploading data to multiple Base Stations

The second network model, as shown in Fig. 1: b, is a non-fully connected scenario in which the sensor nodes upload data to different receiver base stations, while they form an arbitrary mesh topology among themselves. In this case, collisions happen at the receiver base stations and the learning goal is to minimize such collisions in a distributed manner. Unlike in the fully connected scenario, a node in this case can receive transmissions only from its direct neighbors.

## III. NETWORK PROTOCOL MODELING AS MARKOV DECISION PROCESS (MDP)

### A. Markov Decision Process

An MDP is a Markov process [3] in which a system transitions stochastically within a state space $\{S_1, S_2, S_3, \ldots, S_N\}$ as a result of actions taken by an agent in each state. When the agent is in state $S_k$ and takes an action $a_k$, a set of transition probabilities determine the next system state $S_{k+1}$. A reward, indicating a physical benefit, is associated with each such transition. Formally, an MDP can be represented by the tuple $(S, A, T, R)$, where $S$ is the state space, $A$ is the set of all possible actions, $T$ is the state transition probabilities, and $R$ is the reward function. For any system, whose dynamic behavior can be represented by an MDP, there exists an optimal set of actions for each state such that the long-term expected reward can be maximized [4]. Such optimal action sequence is referred to as a solution to the MDP.

### B. Reinforcement Learning (RL)

RL is a class of algorithms for solving an MDP [5] without necessarily requiring an exact mathematical model for the system, as needed by the classical dynamic programming methods. As shown in Fig. 2, in RL, an agent interacts with its environment by taking an action, which causes a state-change. Each action results in a reward that the agent receives from the environment. Q-Learning [6], a model-free and value-based reinforcement learning technique, is used in this work. Using Q-Learning, by taking each possible action in all the states repeatedly, an agent learns to take the best set of actions that represents the optimal MDP solution, which maximizes the expected long-term reward. Each agent maintains a Q-table with entries $Q(s, a)$, which represents the Q-value for action $a$ when taken in state $s$. After an action, the Q-value is updated using the Bellman's equation given by Eq. (1), where $r$ is the reward received, $\alpha$ is a learning rate, $\gamma$ is a discount factor, and $s'$ is the next state caused by action $a$.

$$Q(s,a) \leftarrow Q(s,a) + \alpha[r(s,a) + \gamma \times \max_{\forall a' \in A} Q(s', a') - Q(s,a)] \quad (1)$$

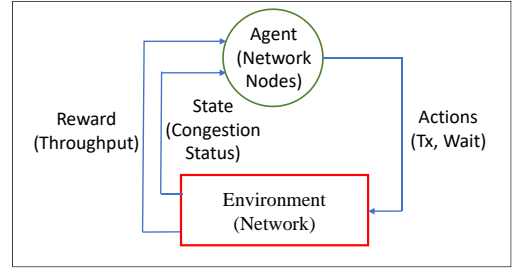

Fig. 2: Reinforcement Learning (RL) for network protocol MDP

### C. Modeling Network Protocols as MDP

We represent the MAC layer protocol logic using an MDP, where the network nodes behave as the learning agents. Environment is the network within which the nodes/agents interact via their actions. The actions here are to transmit with various probabilities and to wait, which the agents need to learn in a network state-specific manner so that the expected reward, which is throughput, can be maximized. A solution to this MDP problem would represent a desirable MAC layer protocol.

<u>State Space</u>: The MDP state space for an agent/node is defined as the network congestion level as perceived by the agent. Congestion is coded with two elements, namely, *inter-collision probability* and *self-collision probability*. An inter-collision takes place when the collided packets at a receiver are from two different nodes. A self-collision occurs at a transmitter when a node's application layer generates a packet in the middle of one of its own ongoing transmissions. It will be shown later as to how using RL, a node learns to deal with both *self-collisions* and *inter-collisions*, thus maximizing the reward or throughput.

Inter-collision and self-collision probabilities are defined by Eqns. (2) and (3) respectively. To keep the state space discrete, these probabilities are discretized in multiple distinct ranges as described in the experimental results sections.

$$P_{SC} = \frac{Number\ of\ self\ collisions}{Number\ of\ transmitted\ packets} \quad (2)$$

$$P_{IC} = \frac{Number\ of\ inter\ collisions}{Number\ of\ transmitted\ packets} \quad (3)$$

<u>Action Space</u>: As for the agents' actions, two different formulations are explored, namely, *fixed actions* and *incremental actions*. With the first one, the actions are defined as the probabilities of packet transmissions by an agent/node. For the latter, the actions are defined as the change of the current transmission probability. Details about these two action space formulations and their performance are presented in Section V.
<u>Rewards</u>: In a fully connected network, each RL agent/node keeps track of its own throughput and those of all other agents. The latter is computed based on the successfully received packets from other nodes. Using such information, an agent-$i$ constructs a reward function as follows.

$$R_i = \{\rho \times S + \sum_{\forall i} \mu_i \times s_i + \sigma \times (f_i)\} \quad (4)$$

$S$ is the network-wide throughput (expressed in packets/packet duration, or Erlang) computed by adding measured throughput for all nodes including node-$i$ itself; $s_i$ is the throughput of node-$i$, and $f_i = -\sum_{\forall j \neq i}|s_i - s_j|$ represents a fairness factor. A larger $f_i$ indicates a fairer system from node-$i$'s perspective. The coefficients $\rho$ and $\sigma$ are learning hyper-parameters. The coefficient $\rho$ provides weightage towards maximizing network-wide throughput, and $\sigma$ contributes towards making the throughput distribution fairer. The node-specific parameter $\mu_i$ determines a media access priority for node-$i$. In addition to the baseline reward structure in Eqn. 4, a learning recovery protection is provided by assigning a penalty of 0.8 to all agents if the network-wide throughput $S$ ever goes to zero.

For a non-fully connected topology, as shown in Fig. 1: b, a modified reward function has been formulated since the individual nodes here do not have the information about the entire network. In this case, a node is able to monitor the number of non-overlapping and overlapping transmissions with only its one hop-neighbor nodes. Let $s_j^i$ be the number of non-overlapping (with that of node-$i$) transmissions per packet duration from node-$j$, as observed by node-$i$. The quantity $s_j^i$ may not represent the actual throughput of node-$j$ since some of those packets may collide with node-$j$'s own neighbors. Similarly, some of node-$j$'s transmissions that are overlapping with node-$i$'s transmissions may actually succeed at one of node-$j$'s neighbor receivers. In the absence of such 2-hop information, we assume that $s_j^i$ represents a reasonable estimate of node-$j$'s throughput from node $i$'s perspective. The reward function for the non-fully connected topology is formulated as:

$$R_i = \rho_i \times (\sum_{\forall i} s_j^i + s_i) + \sigma_i \times f_i \quad (5)$$

Fairness of bandwidth distribution across the nodes within node-$i$'s immediate neighborhood is given by the term $f_i = -\sum_{\forall j \neq i}|s_i - s_j^i|$, where $j$ represents all node-$i$'s one hop neighbors. The coefficient $\rho_i$ accounts for maximizing the estimated throughput of node-$i$ and its one-hop neighbors, whereas $\sigma_i$ tries to distribute the throughput equally among node-$i$'s neighbors. As in the fully connected scenario, a recovery protection is added here too by assigning a fixed penalty if the quantity $s_i$ goes to zero. More about the design of hyper-parameters for both fully connected and non-fully connected scenarios are presented in Section VI.

IV. DISTRIBUTED MULTI-AGENT REINFORCEMENT LEARNING MAC (DRMRL-MAC)

Using the state and action spaces deigned in Section III, we develop an MRL (Multiagent RL) system in which each network node acts as an independent RL agent. Distributed agent behavior give rise to a medium access control protocol that is termed as DMRL-MAC. We used a special flavor of multi-agent RL, known as Hysteretic Q-learning [7].

<u>Hysteretic Q-learning (HQL)</u>: HQL is used in a cooperative and decentralized multi-agent environment, where each agent is treated as an independent learner. An agent, without the knowledge of the actions taken by the other agents in the environment, learns to achieve a coherent joint behavior. The agents learn to converge to an optimal policy without the requirement of explicit communications. The Q-table update rule in Eqn. (1) is modified here as:

$$\delta = r + (\gamma) \times \max_{\forall a' \in A} Q(s', a') - Q(s, a)$$
$$Q(s, a) \leftarrow \begin{cases} Q(s, a) + \alpha \times \delta, & \text{if } \delta \geq 0 \\ Q(s, a) + \beta \times \delta, & \text{else} \end{cases} \quad (6)$$

In a multi-agent environment, the rewards and penalties received by an agent depend not only on its own action, but also on the set of actions taken by all other agents. Even if the action taken by an agent is optimal, still it may receive a penalty as a result of the bad actions taken by the other agents. Therefore, in hysteretic Q-learning, an agent gives less importance to a penalty received for an action that was rewarded in the past. This is taken into consideration by the use of two different learning rates $\alpha$ and $\beta$. This can be seen in the Q-table update rule in Eqn. 6, where $\alpha$ and $\beta$ are the increase and decrease rates of the Q values. The learning rate is selected based on the sign of the parameter $\delta$. The parameter $\delta$ is positive if the actions taken were beneficial for attaining the desired optimum of the system and vice-versa. In order to assign less importance to the penalties, $\beta$ is chosen such that it is always less than $\alpha$. However, the decrease rate $\beta$ is set to non-zero in order to make sure that the hysteretic agents are not totally blind to penalties. In this way, the agents make sure to avoid converging to a sub-optimal equilibrium.

In the proposed DMRL-MAC protocol, each node in the network acts as a hysteretic agent, which is unaware of the actions taken by the other nodes/agents in the network. The actions are evaluated by the rewards assigned using Eqns. (4) and (5). The fact that the Q-table size is independent of number of nodes, makes the protocol scalable with the network size.

V. SINGLE AGENT REINFORCEMENT LEARNING

Before delving into multi-node scenarios, we experiment with the key concepts of RL-based MAC logic with a single network node executing MAC logic for sending data to a base station.

A. Analysis of Self-collision

A self-collision occurs when a node's application layer generates a packet in the middle of one of its own ongoing transmissions. Consider a situation in which an application generates fixed size packets of duration $\tau$ at the rate $g$ Erlang (with Poisson distribution), and hands them to the MAC layer. The MAC layer transmits the packet occupying the channel for the duration $t = 0$ to $t = \tau$. In order for the packet to not trigger any self-collision, no other packet should arrive from the application layer for the duration $[-\tau, \tau]$, the probability of which is $e^{-2g}$. The other scenario when self-collision will be avoided is when one packet arrived and transmitted anytime during the interval $[-2\tau, -\tau]$, and no packet arrived between the end of the transmission of that prior packet and time $t = 0$. The expected probability of this situation to occur is: $(1 - e^{-g}) \times \left(e^{-\frac{g}{2}}\right)$. Combining those two scenarios:

$$P_{no\_self\_collision} = (1 - e^{-g}) \times \left(e^{-\frac{g}{2}}\right) + (e^{-g}) \times (e^{-g})$$

This leads to the single node throughput:

$$s = g \times (e^{-2g} + e^{-g/2} - e^{-3g/2})$$

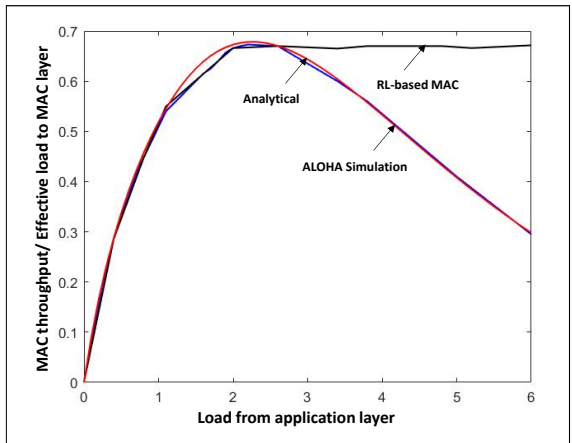

Fig. 3: Single node throughput under self-collisions; throughput comparison between ALOHA and RL-based MAC Logic

Using a *C* language simulator, we have implemented single-node pure ALOHA simulation experiments in which the MAC layer transmits a packet whenever it arrives from the application layer. Fig. 3 shows single-node throughput computed analytically as well for the simulation experiments. The maximum throughput is around 68% at an application layer load of 2.4. In the absence of inter-collisions with other nodes, throughput loss is solely due to the self-collisions modeled above. The throughput reduces with load asymptotically (not shown) as increasing self-collisions eventually prevent any packets to be successfully transmitted.

The analytical model can be further refined by considering the packet arrival history further back compared to just two packet durations. However, the close match between the results from the model and from the experiments in Fig. 3 indicates that such refinements are not necessary.

### B. Handling Self-collisions with Reinforcement Learning

We have programmed a single node to use the classical Q-Learning [2] for medium access. Using the reward formulation in Eqn. 4, the agent learns an appropriate action from the state and action spaces specified in Section III. The state space for single node RL-based MAC is obtained by discretizing the self-collision probability into 6 equal discrete levels. For the fixed action strategy, the actions are defined by the probability of transmission by the node. Five distinct values of transmit probability {0,0.25,0.5,0.75,1} are used. The learning hyperparameters and other relevant Q-Learning parameters are summarized in Table I.

TABLE I: BASELINE EXPERIMENTAL PARAMETERS

| No. of packets per epoch | 1000 |
|---|---|
| $\alpha$ | 0.1 |
| $\gamma$ | 0.95 |
| $\epsilon$ | $0.5 \times e^{-epoch\ ID/200}$ |
| $\rho$ | 1.0 |
| $\sigma$ | 0 |
| $\mu_1$ | 0 |

As shown in Fig. 3, the RL-based MAC is able to achieve the maximum ALOHA throughput by learning to take the appropriate transmission actions from its available action space. However, unlike the regular ALOHA logic, the RL-based logic can maintain the maximum throughout even after the load exceeds 2.4, which is the optimal load for the ALOHA logic. This is achieved by adjusting the transmit probability so that the self-collisions are reduced. For ALOHA, if a node is in the middle of a transmission, it transmits the packets in its queue irrespective of its current transmission status. But with RL, the node can learn not to transmit when it is in the middle of an ongoing transmission. As a result, the effective load $g^*$ to the MAC layer is lowered compared to the original load $g$ from the application layer, thus maintaining the maximum throughout even when the application layer load is increased. Such learning provides a new direction for medium access, which will be explored further for multi-node networks later in the paper.

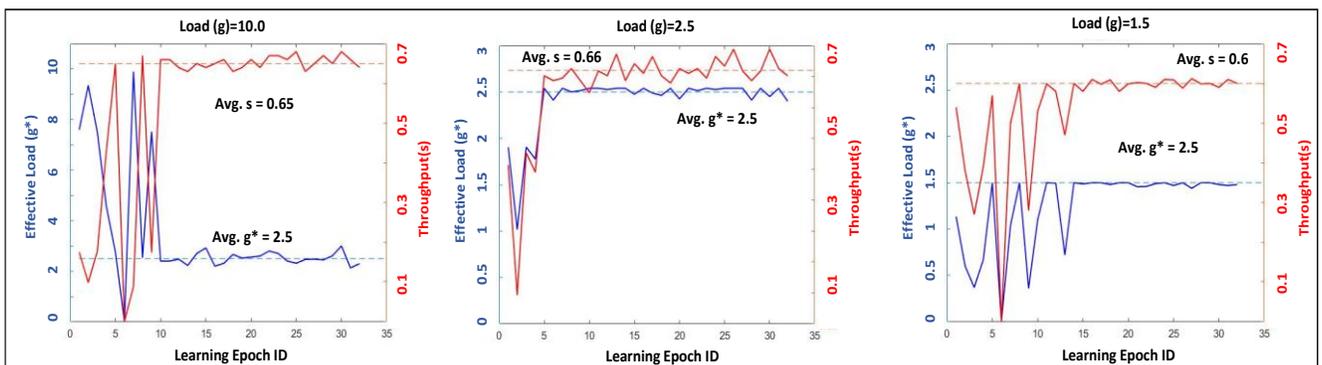

Fig. 4: Convergence plots for RL-based MAC logic: effective load and throughput for three different initial loads using fixed action strategy

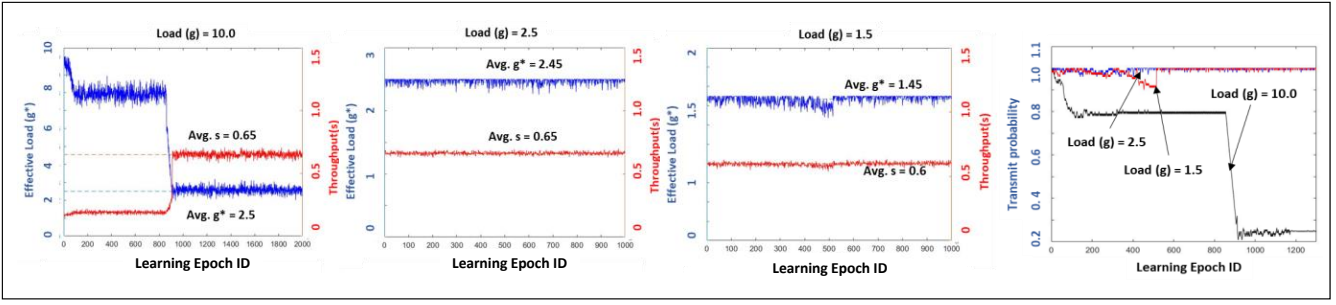

Fig. 5: Convergence plots for RL-based MAC logic: effective load and throughput for three different initial loads using incremental action strategy

As mentioned in Section III, the RL-based MAC is implemented using two different action strategies by the RL agent: fixed action strategy and incremental action strategy. Figs. 4 and 5 show the convergence plots for these two cases respectively. It can be observed that the convergence speed is much higher for the fixed action strategy than that of the incremental one. This is because the search space for the optimal transmit probability is smaller when fixed actions are used as compared to the incremental actions. This is achieved at the expense of accuracy because the granularity of the transmit probability for fixed action strategy is less than that of the incremental action strategy. However, for an application which requires the nodes to learn fast and when the accuracy is not critical, the fixed action strategy proves to be useful.

To summarize, the results in this section for a single-node scenario demonstrates the ability of a reinforcement learning based MAC logic to attain the theoretically maximum throughput and to maintain that for higher application layer loads by controlling self-collisions.

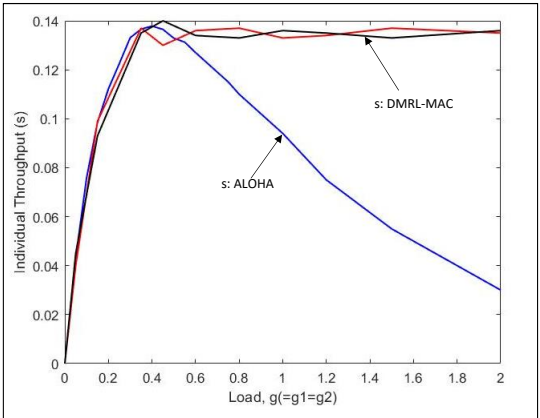

Fig. 6: DMRL-MAC performance in a two-node network: homogeneous load

## VI. MULTI-NODE FULLY CONNECTED NETWORKS

### A. Performance in a Two-Node Network

Unlike for the single node implementation in Section V which uses the classical RL logic, we implemented the Distributed Multi-agent Reinforcement Learning MAC (DMRL-MAC) for multi-node networks. This implementation is based on Hysteretic Q-learning as described in Section IV.

Another key augmentation over the single-node case is that the state space in multi-node scenario contains inter-collision probabilities in addition to the probabilities for self-collisions. In other words, DMRL-MAC uses a 2-dimensional discrete state space with 6 and 4 equal discrete levels of self-collision and inter-collision probabilities respectively.

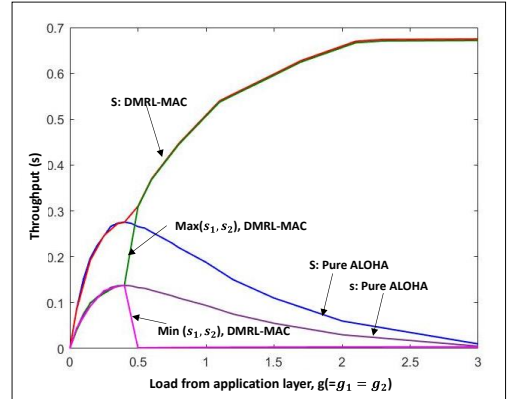

Fig. 7: Performance of DMRL-MAC in a 2-node network with homogeneous load and a total throughput maximization strategy

<u>Homogeneous Loading</u>: As shown in Fig. 6, for the two-node homogeneous loading case, the pure ALOHA protocol can achieve a maximum nodal throughput of $s_1 = s_2 = 0.135$ at the optimal loading $g_1 = g_2 = \hat{g} \approx 0.4$. The figure also shows that the DMRL-MAC logic is able to learn to attain that maximum throughput, and then able to sustain it for larger application layer loads (i.e., $g$). Like in the single-node case, such sustenance is achieved via learning to adjust the effective MAC layer load (i.e., $g^*$) by prudently discarding packets from the application layer. This keeps both self-collisions and inter-collisions at bay for higher throughputs.

We ran DMRL-MAC in a 2-node network with the objective of maximizing networkwide throughput, which is different from maximizing individual throughput in Fig. 7. This was achieved by setting $\mu_i = \sigma = 0$ in Eqn. (4). As shown in Fig. 8, for traffic $g_1 = g_2 < \hat{g}$, the individual node throughputs with DMRL-MAC mimic those of pure-ALOHA. With higher load, however, with DMRL-MAC one of the node's throughput goes to zero so that the other node is able to utilize the entire available throughput, which is the one-node throughput as shown in Fig. 7. This way, the network level throughput is maximized at the expense of the throughput of one of the nodes which is chosen randomly by the underlying distributed reinforcement learning.

<u>Heterogeneous Loading</u>: Results in this section correspond to when the application layer data rates from different nodes are different. Fig. 8 shows the performance for three scenarios, namely, $g_1 < \hat{g}$, $g_2 = \hat{g}$, and $g_1 > \hat{g}$, where $\hat{g}$ is the effective load from the application layer, for which the optimal throughput is obtained for pure ALOHA. For a 2-node network, the value of $\hat{g}$ is found out to be $\approx 0.4$ Erlangs. Node 2's application layer load $g_2$ is varied from 0 to 5 erlangs. The behavior of the system can be categorized in in three broad cases. Case-I: when $g_1 \leq \hat{g}$ and $g_2 \leq \hat{g}$, DMRL-MAC mimics the performance of regular ALOHA. Case-II: when $g_1 \leq \hat{g}$, $g_2 > \hat{g}$ or $g_1 > \hat{g}$, and $g_2 \leq \hat{g}$, the node with the higher load adjusts accordingly such that the optimal ALOHA throughput is maintained. Case III: when $g_1 > \hat{g}$ and $g_2 > \hat{g}$, wireless bandwidth is fairly distributed, and both the nodes transmit such that the effective load boils down to $\hat{g}$. Thus, unlike the regular ALOHA protocol, the DMRL-MAC can learn to maintain the optimal ALOHA throughput for higher traffic scenarios. It does so via learning to reduce both self- and inter-collisions by discarding packets from the application layer.

As can be seen from Fig. 9, an important feature of the RL-based DMRL-MAC protocol is that it can adjust to dynamic traffic environments with time varying loads. When the traffic generated in the network changes, the protocol can adjust transmit probability accordingly, so that the optimal throughput is maintained. It can achieve and maintain the known optimal throughputs and fairly distribute the available bandwidth under heterogeneous loading conditions. This is useful in scenarios in which application layer packet generation fluctuates over time.

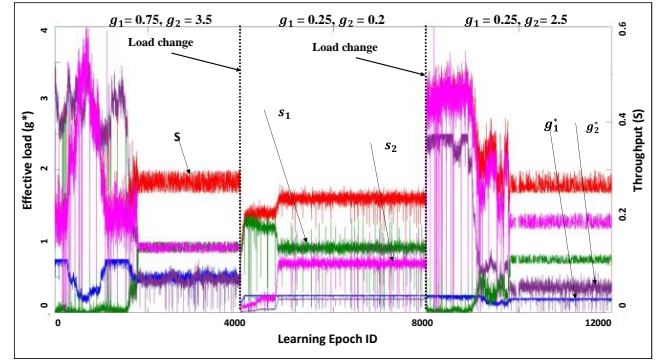

Fig. 9: Convergence plot for DMRL-MAC for dynamic load

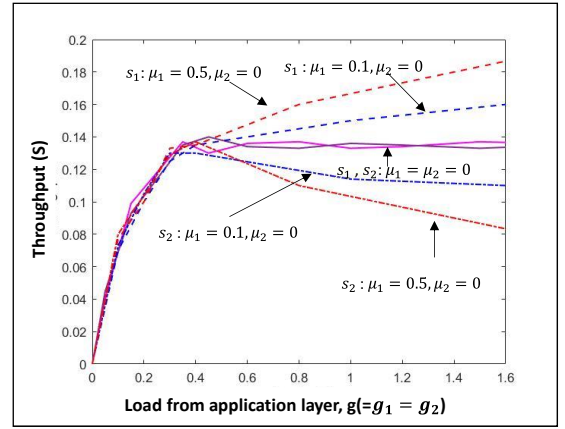

Fig. 10: Load vs throughput plots for different values of $\mu_i$ (priority between the nodes) for a two-node network.

<u>Prioritized Access</u>: One notable feature of the proposed DMRL-MAC is that node-specific access priorities can be achieved by assigning specific values of the coefficients $\mu_i$ in Eqn. (4). In Fig. 10, the load-throughput plots are shown for two different values of $\mu_i$: $\mu_1 = \mu_2 = 0$ and $\mu_1 = 2.0, \mu_2 = 0.1$. For $g_1 \leq \hat{g}$ and $g_2 \leq \hat{g}$, DMRL-MAC mimics the performance of Pure ALOHA for any values of $\mu_i$. If $\mu_1 = \mu_2 = 0$, the system performs as ALOHA, that is, the individual throughput for each node is equal to the ALOHA maximum. With increase in $\mu_1$,

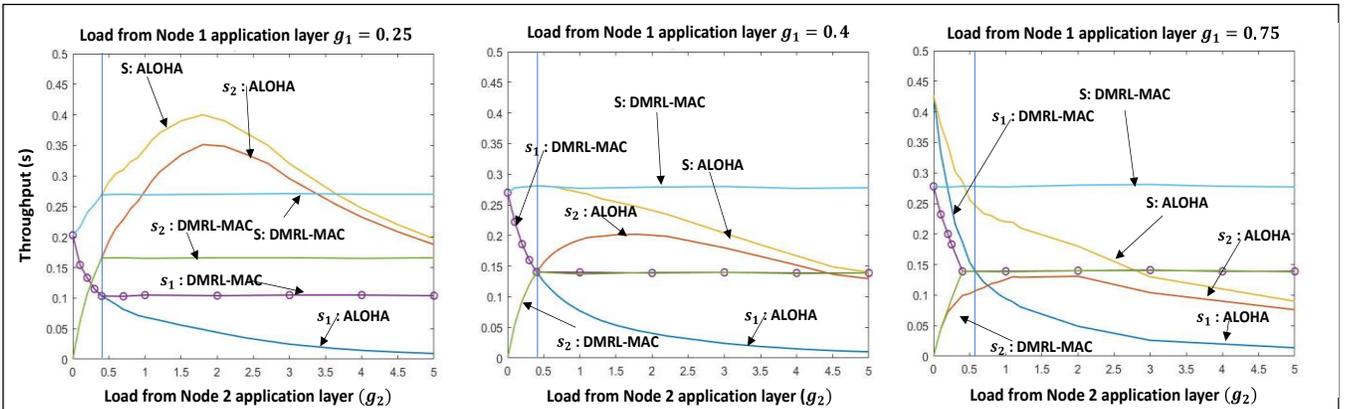

Fig. 8: Performance of DMRL-MAC in a two-node network with heterogeneous load [$s_1, s_2$ are individual throughputs of node 1 and node 2 respectively; S is the networkwide throughput; $g_1, g_2$ are load (in Erlangs) from the application layers of node 1 and node 2 respectively]

node 1 gets the priority and the individual throughput for node 2 approaches towards zero. This kind of prioritized access is useful when data from specific sensors are more critical compared to others, especially when the available wireless bandwidth is not sufficient.

### B. Performance in Larger Networks

Performance of DMRL-MAC for 3-node network is shown in Fig. 11. As shown for the simulated ALOHA performance, the maximum network wide throughput for a homogeneous load distribution occurs when $g_1 = g_2 = g_3 = \hat{g} \approx 0.25$ erlangs. That throughput is $S = 0.26$, and that is with a fair distribution among the nodes. It can be observed that like the 1-node and 2-node scenarios, DMRC-MAC can learn the theoretically feasible maximum throughput and maintain that at higher loads by avoiding both self- and inter-collisions.

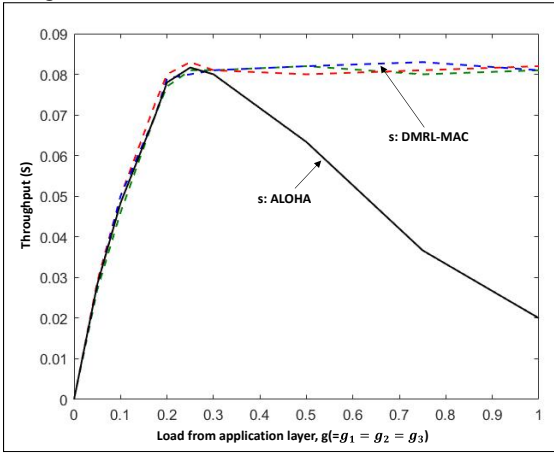

Fig. 11: Performance of DMRL-MAC in a three-node network with homogeneous load

For large networks with 2 or higher node-count, the learning hyperparameters in Eqn. (4) are made empirically dependent on the network size $N$ as follows. We set $\sigma = \frac{1}{N-1}$ because with larger $N$, both the number of contributing terms in the expression of $f_i$ in Eqn. (4) and the value of $f_i$ itself go up. This effect is compensated by making $\sigma$ (the coefficient of $f_i$) inversely proportional to the number of one-hop neighbors (which is *N-1* for a fully connected network). After setting the value of $\sigma$, the parameter $\rho$ in Eqn. (4) is determined empirically. It is observed that for a given $\sigma$, a range of values of $\rho$ can be obtained for which the system converges. That range decreases with larger *N*. The relationship was experimentally found as: $\rho = 0.33 - 0.05 \times N$. Using this empirical relationship, the reward expression from Eqn. 4 can be rewritten as:

$$R_i = \{(0.33 - 0.05 \times N) \times S + \sum_{\forall i} \mu_i \times s_i + \frac{f_i}{N-1}\} \quad (5)$$

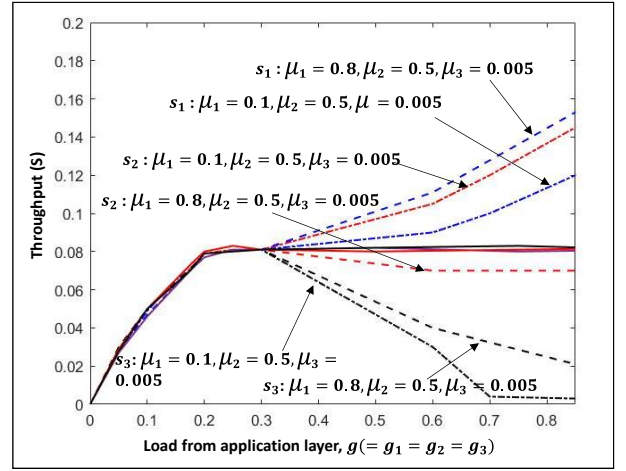

Fig. 13: Load versus throughput plots for different values of $\mu_i$ (priority among the nodes) for a three-node network.

Performance under heterogeneous loads is analyzed and reported in Fig. 14. Three different situations are studied, namely, $g_1 \leq \hat{g}, g_2 \leq \hat{g}$, $g_1 \leq \hat{g}, g_2 > \hat{g}$ or $g_1 > \hat{g}, g_2 \leq \hat{g}$, and $g_1 > \hat{g}, g_2 > \hat{g}$. In all these three cases, throughput variation is studied by varying $g_3$. It can be seen that for $g_1 \leq \hat{g}, g_2 \leq \hat{g}$, and $g_3 \leq \hat{g}$, DMRL-MAC mimics the performance of regular ALOHA. When the load in any of these nodes goes higher than the optimal value ($\hat{g}$), learning enables the node to adjust transmit probability so that the optimal ALOHA throughput is maintained by limiting both types of collisions.

Priority among the nodes can be assigned using the coefficient $\mu_i$ in the reward function in Eqn. (4). As shown in

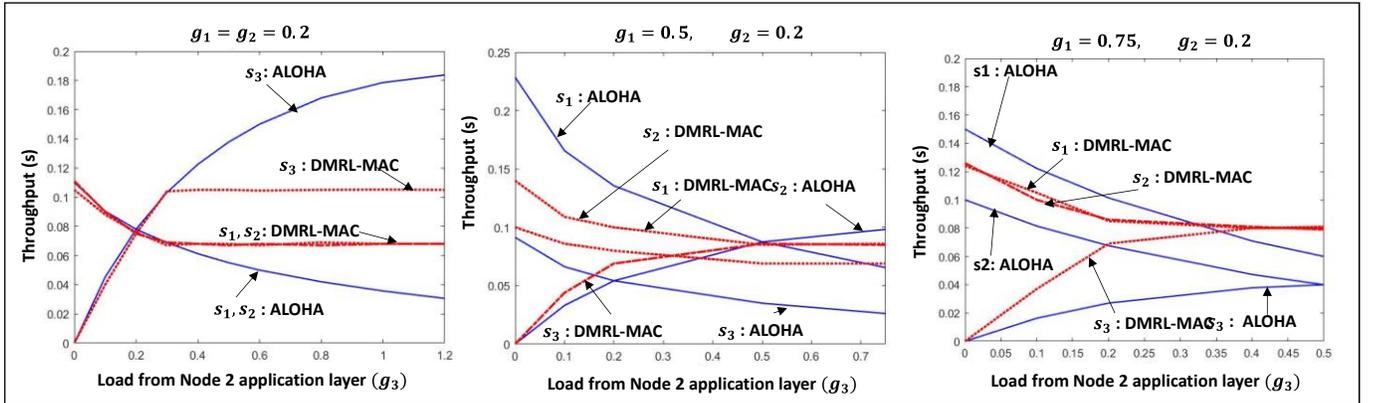

Fig. 12: Performance of DMRL-MAC in a three-node network with heterogeneous load [$s_1, s_2, s_3$ are individual throughputs of node 1, node 2 and node 3 respectively; S is the networkwide throughput; $g_1, g_2, g_3$ are load (in Erlangs) from the application layers of node 1, node 2 and node 3 respectively]

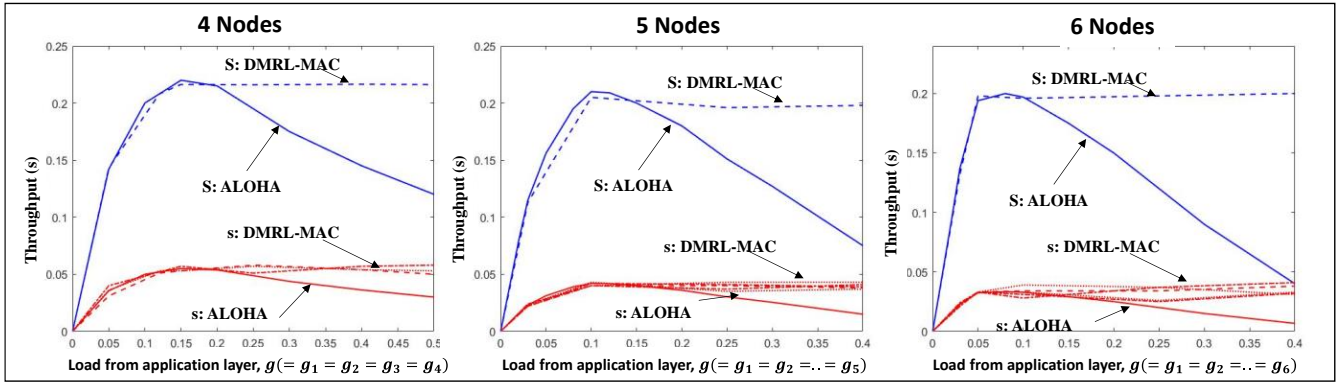

Fig. 14: Performance of DMRL-MAC in a network with 4,5 and 6 nodes for homogeneous load

Fig. 13., when $g_1 \leq \hat{g}$ and $g_2 \leq \hat{g}$, DMRL-MAC mimics the performance of Pure ALOHA for any values of $\mu_i$. Also, if $\mu_1 = \mu_2 = \mu_3 = 0$, the channel bandwidth is fairly distributed. However, when the values of $\mu_2$ and $\mu_3$ are set to 0.5 and 0.005 respectively, with an increase in $\mu_1$, the throughput for node 1 (i.e., $s_1$) increases. The node with the largest $\mu_i$ value gets the highest portion of the available wireless bandwidth. These results demonstrate how DMRL-MAC's ability of prioritized access can hold for a 3-node fully connected networks.

Fig. 14 depicts the performance of larger networks with 4, 5, and 6 nodes. It shows that the desirable properties of DMRL-MAC in attaining the theoretical maximum throughput and maintaining it at higher loads are still valid for such larger networks. However, the convergence becomes increasingly slower as the networks grow in size. The convergence time distributions in Fig. 15 in fact show that the learning almost stops working for networks with 7 or more nodes.

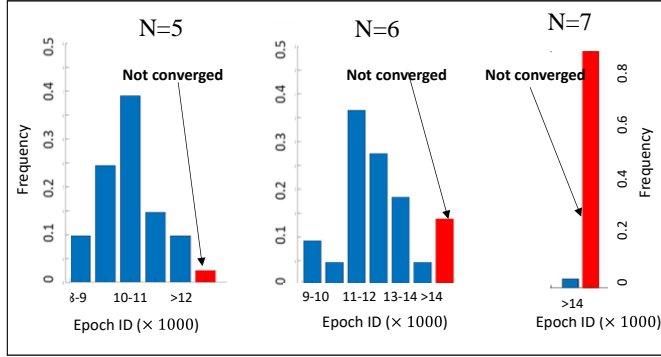

Fig. 15: Convergence behavior for different number of nodes (pmf)

To investigate this more, we have plotted the self- and inter-collisions probabilities in Fig. 16. The first notable observation is that while the self-collisions increases with higher application layer loads, it is almost insensitive to the network size. As expected, the inter-collisions do increase with larger network size. However, the rate of increase of the inter-collisions with increasing application layer loads reduces in larger networks. The implication of this observation is that as the network size increases, the inter-collisions become less and less sensitive to individual node's transmission decisions. In other words, the reinforcement learning agents' actions become less influential for changing the system's states. Thus, the system becomes stateless and moves out of the realm of reinforcement learning based solution discussed here. A stateless Q-learning or Multi-arm bandit-based MDP solution [2] is needed for this situation, which will be explored as the next step beyond this paper, which reports our initial approaches and preliminary results.

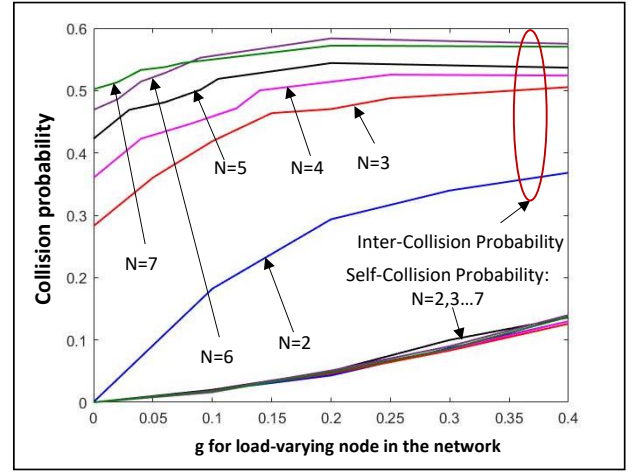

Fig. 16: Variation of inter-collision and self-collision probability with number of nodes (N) and traffic load (g)

## VII. PARTIALLY CONNECTED MULTI-NODE NETWORKS

Although the primary goal of this paper is to report the key concepts of RL based wireless medium access and its performance in fully connected networks, here we include a preliminary set of results to demonstrate how the protocol behaves in a partially connected scenario shown in Fig. 1(b). In this network, since *node-2* can listen to both *node-1* and *node-3*, the transmissions from *node-2* experience more collisions than those experienced by *node-1* and *node-3*. Fig. 17(a) shows the load-throughput plot for this network when the nodes run regular ALOHA protocol. It is observed that the maximum throughput for *node-2* ($s_2^{max} = 0.078$) is half the maximum throughput for *node-1* and *node-3* ($s_1^{max} = s_3^{max} = 0.155$). The goal of the proposed DMRL-MAC is to adjust the transmit probability of each node such that the channel bandwidth is maximized, with the condition that it is fairly distributed among the three nodes. Fig. 17 (b) shows the throughput variation with the load $g_1(= g_3)$ when the throughput is equal for all three nodes. It can be seen from the plot that due to more collisions of packets from *node-2*, the value of $g_2$ needs to be larger than $g_1$ and $g_3$ for receiving equal bandwidth. This analysis provides the benchmark for testing the DMRL-MAC protocol: $\widehat{g_1} = \widehat{g_3} = 0.2, \widehat{g_2} = 0.28$ and $\hat{S} = 0.31$. This benchmark throughput is lower than the maximum attainable throughput ($S^{max} = 0.37$) with homogeneous load.

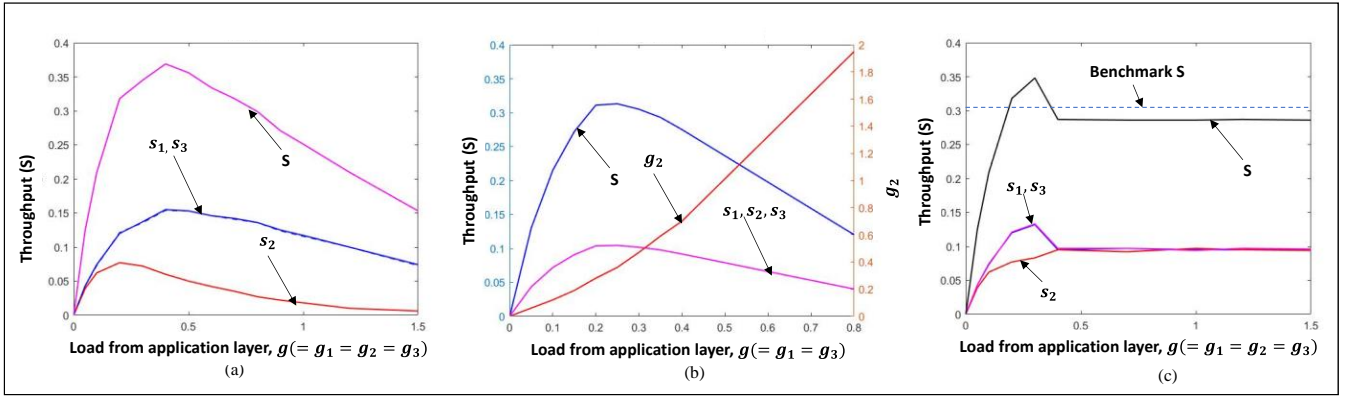

Fig. 17: Load vs throughput plot (a) for ALOHA when $g_1 = g_2 = g_3$, (b) for ALOHA when $s_1 = s_2 = s_3$, (c) for DMRL-MAC when $g_1 = g_2 = g_3$

Implementation of DMRL-MAC for a partially connected network is different from the fully connected case in the following ways. A node in this case have no throughput information about its 2-hop nodes and beyond and have only partial information about the throughput of 1-hop neighbors. A node can monitor the number of transmissions from its 1-hop neighbors that are overlapping and non-overlapping with its own transmissions. In the absence of any network-wide information, a node running DMRL-MAC treats: i) its immediate neighborhood (i.e., 1-hop) as the complete network, and ii) the estimated throughput of its 1-hop neighbors computed from monitored non-overlapping transmissions as their approximated actual throughputs.

Performance of DMRL-MAC executed with such partial information for the partially connected linear network from Fig. 1(b) is shown in Fig. 17(c). It can be observed that this distributed learning-based approach can achieve a near-optimal throughput in this case. In this scenario, when $g_1, g_2, g_3 <$ optimal $\hat{g}$, the system achieves pure ALOHA performance, and for $g_1, g_2, g_3 >$ optimal $\hat{g}$, a near-optimal throughput ($S' = 0.27 (< \hat{S})$) is maintained even at higher loads. Moreover, the throughput is equally distributed among all the nodes even though node-2 is in a more disadvantageous position than node-1 and node-3 in terms of collisions. It should be noted that the results in this section are a preliminary reporting for the partially connected scenarios. More protocol refinements and experiments will be needed to address learning in the absence of complete throughput information in the neighborhood. These results are just to indicate the promise of distributed reinforcement learning in partially connected networks. More results on this will be published elsewhere.

## VIII. RELATED WORK

A number of Reinforcement Learning based network access strategies were explored in the literature. The approaches in [8, 9] apply Q-learning in order to increase MAC layer throughput by reducing collisions in a slotted access system. Unlike the DMRL framework in this paper, the approach in [9] does not solve the access solution in scenarios with load and network topology heterogeneity. The ability to deal with load heterogeneity is also missing in [10,11], which present Q-learning based adaptive learning for underwater networks.

A number of papers explored RL as a means to control network resource allocation. Throughput maximization for secondary nodes in a cognitive network was considered in [12]. In a time-slotted system, under the assumption of a homogeneous secondary network, Q-learning is used for solving a Partially Observable Markov Process (POMDP) in order to reduce the interference between the primary and secondary users. In [13] and [14], Q-learning was explored to find efficient and load-dependent radio transmission schedules. It was shown that the approach delivers good throughput and delay compared to the classical MAC protocols. However, the protocols are designed considering a homogeneous and dense network. Moreover, unlike the proposed DMRL in this paper, the approach in [13] does not learn to do load modulation for maintaining good throughput at higher loads. Two more Q-learning-based resource allocation protocols were proposed in [15] and [16]. These protocols rely on the ability of carrier sensing to dynamically modulate an access contention window [16] so that collisions are reduced. This work is not directly comparable to the work presented here due to its ability of carrier sensing. In [17], the authors develop a RL-based mechanism for scheduling sleep and active transmission periods in a wireless network. While this technique was shown to be able to achieve throughputs that are higher than the traditional MAC protocols, the throughput falls with the increase in the network traffic. Such loss of performance at high application layer load is specifically avoided in our proposed DMRL work.

## IX. SUMMARY AND CONCLUSIONS

A multi-agent Reinforcement Learning based protocol for MAC layer radio access has been proposed. The protocol allows network nodes to individually adjust transmit probabilities so that self-collisions and inter-collisions are reduced. As a result, the nodes can attain the theoretical maximum throughput and sustain it for higher loading conditions. An important feature of the protocol is that it can deal with network heterogeneity in terms load, topology, and QoS needs in the form of access priorities. Moreover, learning allows the nodes to self-adjust in a dynamic environment with a time-varying traffic load. The proposed protocol has been tested for various size networks with nodes without carrier sensing abilities. From a Markov Decision Process (MDP) perspective, it is shown that the system becomes increasingly stateless for larger networks. In such scenarios, stateless Q-learning or Multi-arm bandit-based solutions can be used, which will be explored in a separate paper. As next steps, we will explore MDP solutions in networks with arbitrary topology. We will also explore the framework for nodes with channel sensing capabilities and compare them with existing CSMA family of protocols. Future work will also consider other access performance such as energy and delay.